\documentclass[sigconf]{acmart}

\usepackage{booktabs} 
\usepackage{multirow}
\usepackage{multicol}
\usepackage{subfigure}
\usepackage{graphicx}
\usepackage{verbatim}
\usepackage{balance}

\setcopyright{acmcopyright}

\acmConference[MM'18]{ACM Multimedia}{Oct 2018}{Seoul, Korea}
\acmYear{2018}
\copyrightyear{2018}

\copyrightyear{2018} 
\acmYear{2018} 
\setcopyright{acmcopyright}
\acmConference[MM '18]{2018 ACM Multimedia Conference}{October 22--26, 2018}{Seoul, Republic of Korea}
\acmBooktitle{MM '18: 2018 ACM Multimedia Conference, Oct. 22--26, 2018, Seoul, Republic of Korea}
\acmPrice{15.00}
\acmDOI{10.1145/3240508.3240702}
\acmISBN{978-1-4503-5665-7/18/10} 

\begin{document}
\title{PVNet: A Joint Convolutional Network of Point Cloud and Multi-View for 3D Shape Recognition}
  
\author{Haoxuan You}
\affiliation{%
  \institution{Tsinghua University}
  \city{Beijing}
  \state{China}
}
\email{haoxuanyou@gmail.com}

\author{Yifan Feng}
\affiliation{%
  \institution{Xiamen University}
  \city{Xiamen}
  \state{China}
}
\email{evanfeng97@gmail.com}

\author{Rongrong Ji}
\affiliation{%
  \institution{Xiamen University}
  \city{Xiamen}
  \state{China}
}
\email{rrji@xmu.edu.cn}

\author{Yue Gao}
\authornote{Corresponding author.}
\affiliation{%
  \institution{Tsinghua University}
  \city{Beijing}
  \state{China}
}
\email{kevin.gaoy@gmail.com}




\renewcommand{\shortauthors}{Haoxuan You, Yifan Feng, Rongrong Ji, Yue Gao.}

\begin{abstract}
3D object recognition has attracted wide research attention in the field of multimedia and computer vision. With the recent proliferation of deep learning, various deep models with different representations have achieved the state-of-the-art performance. Among them, point cloud and multi-view based 3D shape representations are promising recently, and their corresponding deep models have shown significant performance on 3D shape recognition. However, there is little effort concentrating point cloud data and multi-view data for 3D shape representation, which is, in our consideration, beneficial and compensated to each other. In this paper, we propose the Point-View Network (PVNet), the first framework integrating both the point cloud and the multi-view data towards joint 3D shape recognition. More specifically, an embedding attention fusion scheme is proposed that could employ high-level features from the multi-view data to model the intrinsic correlation and discriminability of different structure features from the point cloud data. In particular, the discriminative descriptions are quantified and leveraged as the soft attention mask to further refine the structure feature of the 3D shape. We have evaluated the proposed method on the ModelNet40 dataset for 3D shape classification and retrieval tasks. Experimental results and comparisons with state-of-the-art methods demonstrate that our framework can achieve superior performance.
\end{abstract}



\maketitle

\section{Introduction}
3D  data recognition and analysis is surely a fundamental and intriguing area in multimedia and computer vision, spanning broad applications from environment understanding to self-driving. How to understand 3D data, such as recognizing 3D shapes, has attracted much attention in recent years. With the development of deep learning, various deep networks have been employed to deal with different kinds of 3D data, point clouds, multi-view and volumetric data. While it is natural and reasonable to extend 2D convolutional neural networks to volumetric data \cite{wu20153d, maturana2015voxnet, qi2016volumetric}, these methods suffer from the large computational complexity and data sparsity , making it difficult to deal with high image resolution. In contrast, analyzing either the multi-view data \cite{su2015multi} or the point cloud data \cite{qi2017pointnet, qi2017pointnet++,klokov2017escape} have been more flexible and reported better performance, due to their wider source of data acquisition and storage. Therefore, recent works mainly concentrate on either point cloud based or multi-view based 3D shape recognition.

\begin{figure}
\begin{center}
\includegraphics[width=3.4in]{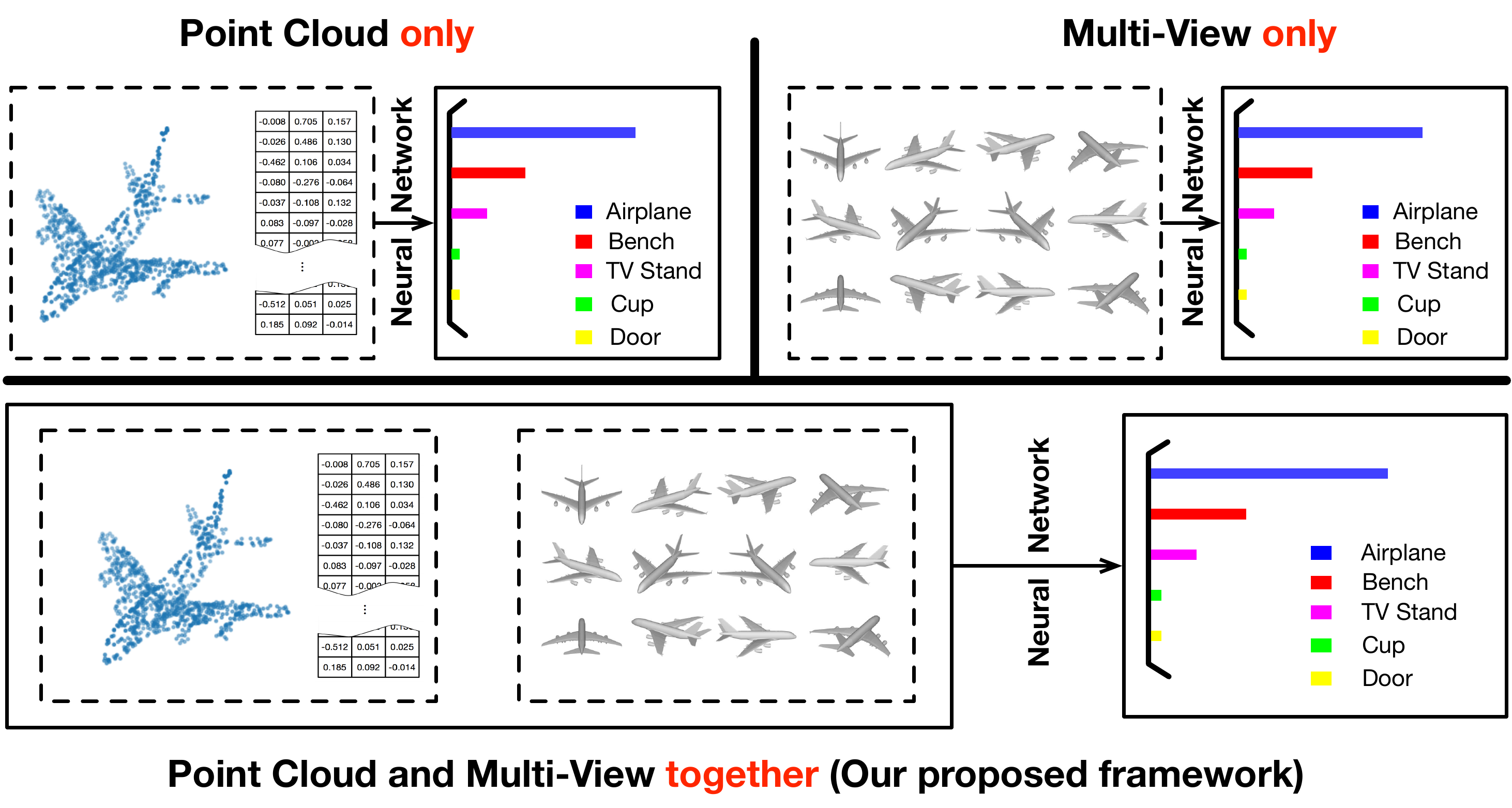}
\end{center}
\caption{Top: Illustration of the 3D shape recognition framework based on point cloud and multi-view. Bottom: Illustration of our proposed joint framework using point cloud and multi-view together.}
\label{Fig:overview}
\end{figure}

\begin{figure*}[htbp]
\begin{center}
\includegraphics[width=1\textwidth]{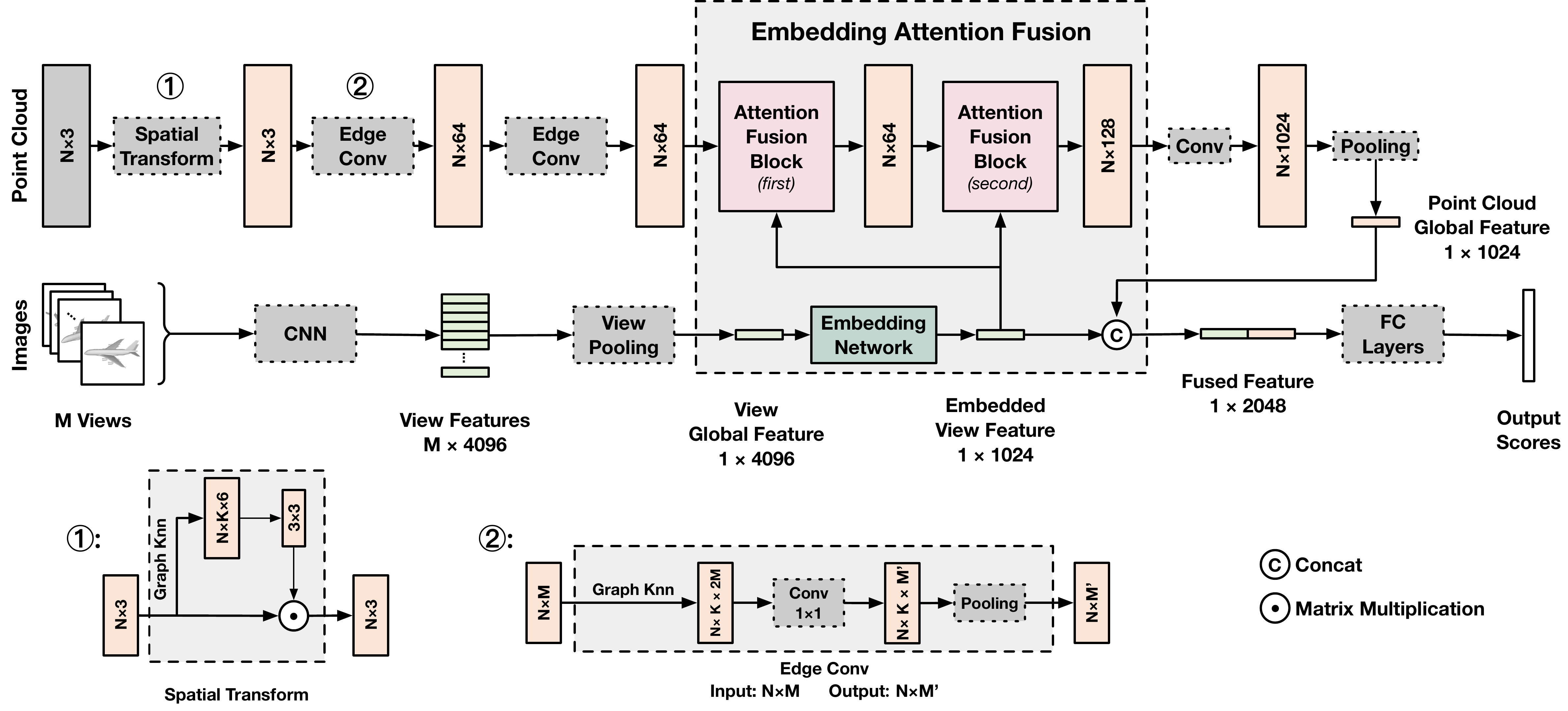}
\end{center}
\caption{Our PVNet framework is composed of 3 parts: point cloud branch, multi-view branch and the embedding attention fusion. Point cloud branch: This branch takes $n$ points with 3-dimensional coordinates as input. Then in spatial transform net, a $3 \times 3$ matrix is learned to align the input points to a canonical space. For EdgeConv, it extracts the local patches of each point by their $k$-nearest neighborhoods and computes edge features for each point by applying a $1 \times 1$ convolution with output channels $M'$, and then generates the tensor after pooling among neighboring edge features. Multi-view branch: The structure of MVCNN is employed, which contains a weight-shared CNN and following view pooling layer that conducts max pooling across all views. Embedding attention fusion: The global view feature is projected into the subspace of point cloud feature by an embedding network. Besides directly combined with point cloud global feature, the embedded view feature is efficiently leveraged in attention fusion block to generate more discriminative features.}
\label{fig:pipline}
\end{figure*}

In multi-view based methods, each 3D shape is represented by a group of views generated from different directions. In this way, the input data can be easily processed by exploiting different well-established convolutional neural networks (CNN), such as AlexNet \cite{krizhevsky2012imagenet}, VGG \cite{simonyan2014very}, GoogLeNet \cite{szegedy2015going}, ResNet \cite{he2016deep}, and DenseNet \cite{huang2017densely}. For example, Su et al. \cite{su2015multi} introduced a Multi-View Convolutional Neural Networks for 3D shape recognition, in which the view-level features were used to generate the shape-level feature by pooling. Qi et al. \cite{qi2016volumetric} proposed to employ multi-resolution views for shape representation. Although the high-level view features can be efficiently extracted with good performance on different tasks, the multi-view based representation, influenced by the camera angles, inevitably discards the information of local structures.

In point cloud based methods, the point cloud data can be obtained from the raw output of most 3D sensing devices, which can better preserve the 3D spatial information and internal local structure. But Point cloud data are usually disordered and irregular, and rendering conventional 2D CNN is unsuitable to exploit the structure correlation from the point cloud data. Various models are designed to handle the extraction of structure features in point clouds. As a pioneer work, PointNet \cite{qi2017pointnet} learns the spatial feature of each point independently and then accumulates the features by a symmetric function. Furthermore, subsequent models \cite{qi2017pointnet++,wang2018dynamic,shen2017neighbors} have proven that considering the neighborhoods of points rather than treating points independently can better leverage the local structure features to improve the performance. We note that these methods mainly focus on how to extract local features, and different local structures should contribute differently for 3D shape representation. However, the relationships among different local structure features are still left unexplored in existing methods that employ point cloud data only, which limits the representation ability of point cloud for 3D shapes.

In view of the prob./neg. of the existing point cloud based methods and multi-view based methods, it is a natural thought, it is necessary to employ high-level global features from the multi-view data to mine the relative correlations between different local features from the point cloud data. To tackle the issue, in this paper, we propose a point-view network (termed  PVNet) , which can serve as the first attempt towards joint point cloud and multi-view based 3D shape representation. In PVNet, an attention embedding fusion is introduced to combine the two types of 3D data. First, the global features taken from the multi-view data are projected into the subspace of point cloud features by an embedding network. Second, these embedded high-level global features are exploited in a proposed attention mechanism to help to learn the relationships between local features extracted from the point cloud data. More specifically, soft attention masks can be adaptively generated by fusing the embedded global features and the point cloud local features, which describes the significance of different local structures. Third, the masks are applied to the point cloud based network in a residual way to enhance more discriminative features and suppress useless features. Besides, the embedded features are also fused at the last fully-connected layers. To evaluate the performance of our proposed PVNet framework, we have conducted experiments on the ModelNet40 dataset, with comparisons to the state-of-the-art methods, including both multi-view based and point cloud based models. Experimental results show that our proposed PVNet method can achieve better performance on both 3D shape classification and retrieval tasks, which demonstrates the effectiveness of the proposed framework. 

The main contributions of this paper are two-fold;
\begin{enumerate}
\item We propose the first convolutional network , i.e., PVNet, that jointly considers both the point cloud data and the multi-view data for 3D shape recognition. Different from conventional 3D deep models, our framework employs the high-level global features from the multi-view data to complement the strengths of feature extraction of the point cloud data. Our framework is compatible with different view and point cloud models.
\item We design an attention embedding fusion method. From the embedded global features of view models, our method can obtain soft attention masks adaptively to generate the attention-aware features of point cloud models, which are more efficient in representing discriminative information of 3D data.
\end{enumerate}

The rest of this paper is organized as follows. We first introduce the related work in Sec.2. We then present the proposed PVNet framework in Sec.3. Experiments and discussions are provided in Sec.4. Finally, we conclude this paper in Sec.5.

\section{Related Work}
In this section, we briefly review existing works of 3D shape classification and retrieval from view based models, point cloud based models and multimodal fusion methods.
\subsection{View Based Models}
View based models take a group of views captured from different angles to describe the 3D object. Hand craft descriptors are investigated in the beginning. Lighting Field descriptor \cite{chen2003visual}, the first typical view based 3D descriptor, is composed of a group of ten views and captured from the vertices of a dodecahedron over a hemisphere. The probabilistic matching is employed to measure the similarity between two 3D objects in \cite{gao2012camera}. With the development of deep learning, many models based on deep neural networks are widely investigated. A multi-view convolutional neural network is proposed in \cite{su2015multi}, which first generates the features of each view by a weight-shared CNN, followed by a view pooling to aggregate them. In \cite{feng2018gvcnn}, a group-based method is proposed to exploit the relationship among different views. For the retrieval task, a low-level Mahalanobis metric is applied to boost the performance. In \cite{guo2016multi},  a deep embedding network is employed in a unified multi-view 3D shape retrieval method to solve the complex intra-class and inter-class variations, in which the deep convolutional network can be jointly supervised by classification loss and triplet loss. A deep auto-encoder structure is also proposed in \cite{xie2017deepshape} to better extract shape features.

\begin{figure}
\begin{center}
\includegraphics[width=2.4in]{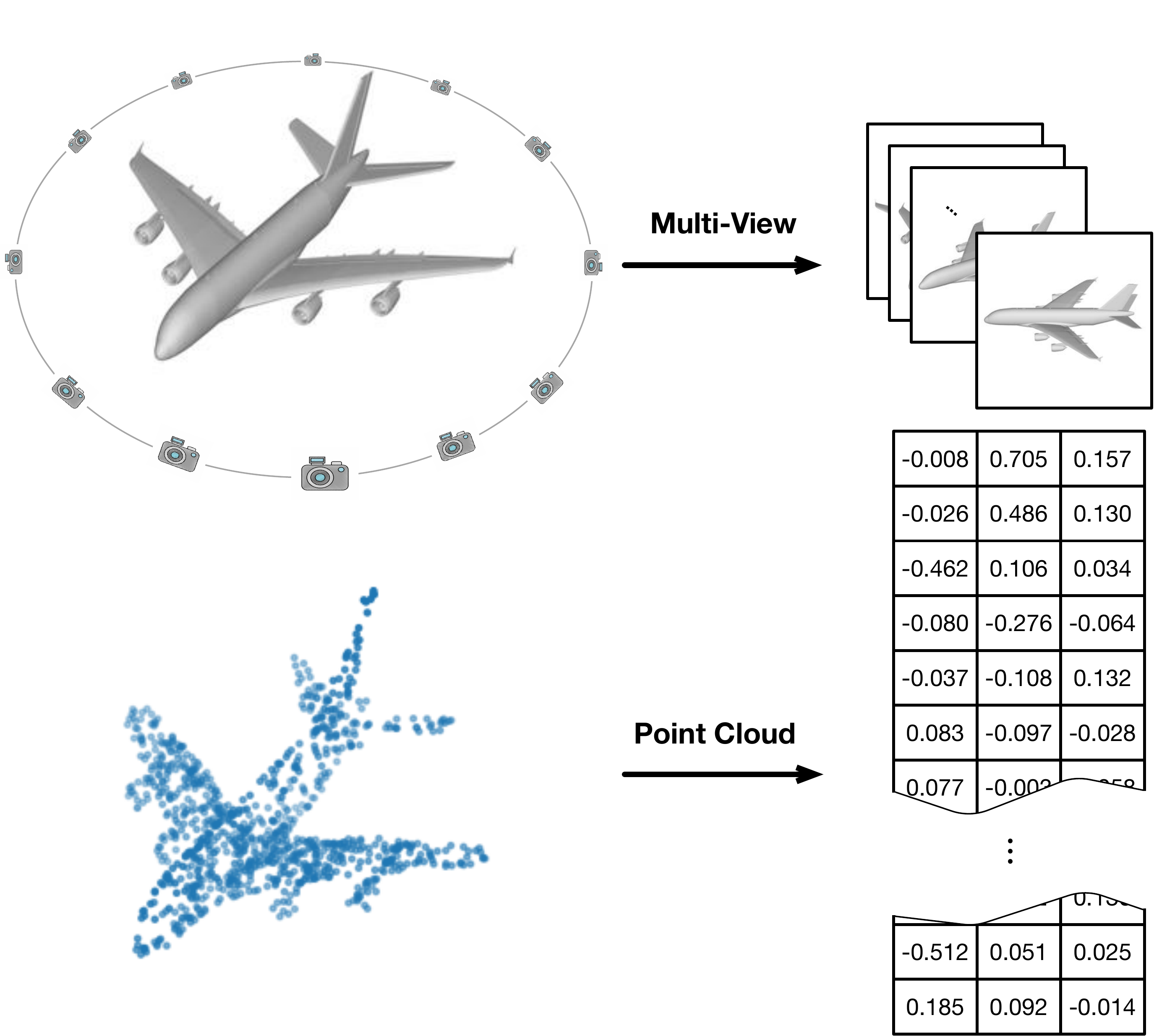}
\end{center}
\caption{Input data for our PVNet. Top: The views are generated by the predefined camera array, and we use 12 views for training our network. Bottom: A point cloud set contains $n$ points with their 3D coordinates.}
\label{fig:input_data}
\end{figure}

\subsection{Point Cloud Based Models}
Point cloud is a straightforward representation to describe naive 3D object. PointNet \cite{qi2017pointnet} first introduces deep neural networks to directly process point clouds. Spatial transform network and a symmetry function are used for the invariance to permutation. Even though PointNet could achieve exciting performance in classification and segmentation tasks, the local geometric features can not be well extracted for treating point independently. 
The recent works mainly focus on how to efficiently capture local features. For instance, PointNet++ \cite{qi2017pointnet++} applies PointNet structure in local point sets with different resolutions and accumulates local features in a hierarchical architecture. Kd-Network \cite{klokov2017escape} extracts and aggregates features according to the subdivisions of the point clouds on Kd-trees. In \cite{wang2018dynamic}, EdgeConv is proposed as a basic block to build networks, in which the edge features between points and their neighbors are exploited. PointCNN \cite{li2018pointcnn} proposes $\chi$-Conv to aggregate features in each local pitches and applies a hierarchical network structure likes typical CNNs. However, existing models still have not exploited the correlations of different geometric features and their discriminative information toward final results, which limits the performance.

\subsection{Multimodal Fusion Methods}
For 3D object, only one kind of representation would hardly cover all the properties and different representations have distinct models to exploit features. it is therefore beneficial to fuse different models in an efficient and reasonable way for better leveraging their respective advantages. In \cite{hegde2016fusionnet}, the Volumetric CNN architecture is proposed to combine volumetric and view representation together to learn new features. In the context of autonomous driving, images, depth and optical flow are fused together by a mixture-of-experts framework for 2D pedestrian detection \cite{gonzalez2017board}. MV3D network \cite{chen2017multi} is proposed to fuse point cloud data from LIDAR and view data, which however projects the point cloud data into the form of images and processes them together with view data by CNNs. For attentive multimodal fusion, in \cite{poria2017multi}, textual, audio and visual features are fused in an attention-based recurrent model. In \cite{gao2018question}, the question-guided hybrid convolution with attention mechanism is proposed to couple the textual and visual information in VQA system. To sum up, our PVNet is the first framework that directly fuses the point cloud stream and multi-view stream in a sophisticated way. 

\section{Point-View Network}
In this section, we give a detailed introduction to our PVNet framework. Our input is two modalities for 3D shape representation: raw point clouds and 2D views. The point clouds are the subsets of unordered points from a Euclidean space. The 2D views are rendered images of 3D shapes captured by predefined camera arrays. Following the camera setting of \cite{su2015multi}, we employ 1,024 points and 12 views to represent one 3D shape \footnote{we have no limit on other methods to obtain different views.}.  The corresponding input toward a 3D object is shown in Fig. \ref{fig:input_data}.

Given two kinds of input, we feed them into two branches individually: point cloud branch and multi-view branch, which are consisted of basic models well investigated. Then the high-level global features from multi-view branch are incorporated into point cloud branch by the proposed embedding attention fusion, which provides guidance to better describe the significance of different structure features. The final feature after fusion is used both for classification and retrieval tasks. Fig. \ref{fig:pipline} illustrates the detailed flowchart of our framework.

\begin{figure}
\begin{center}
\includegraphics[width=3.2in]{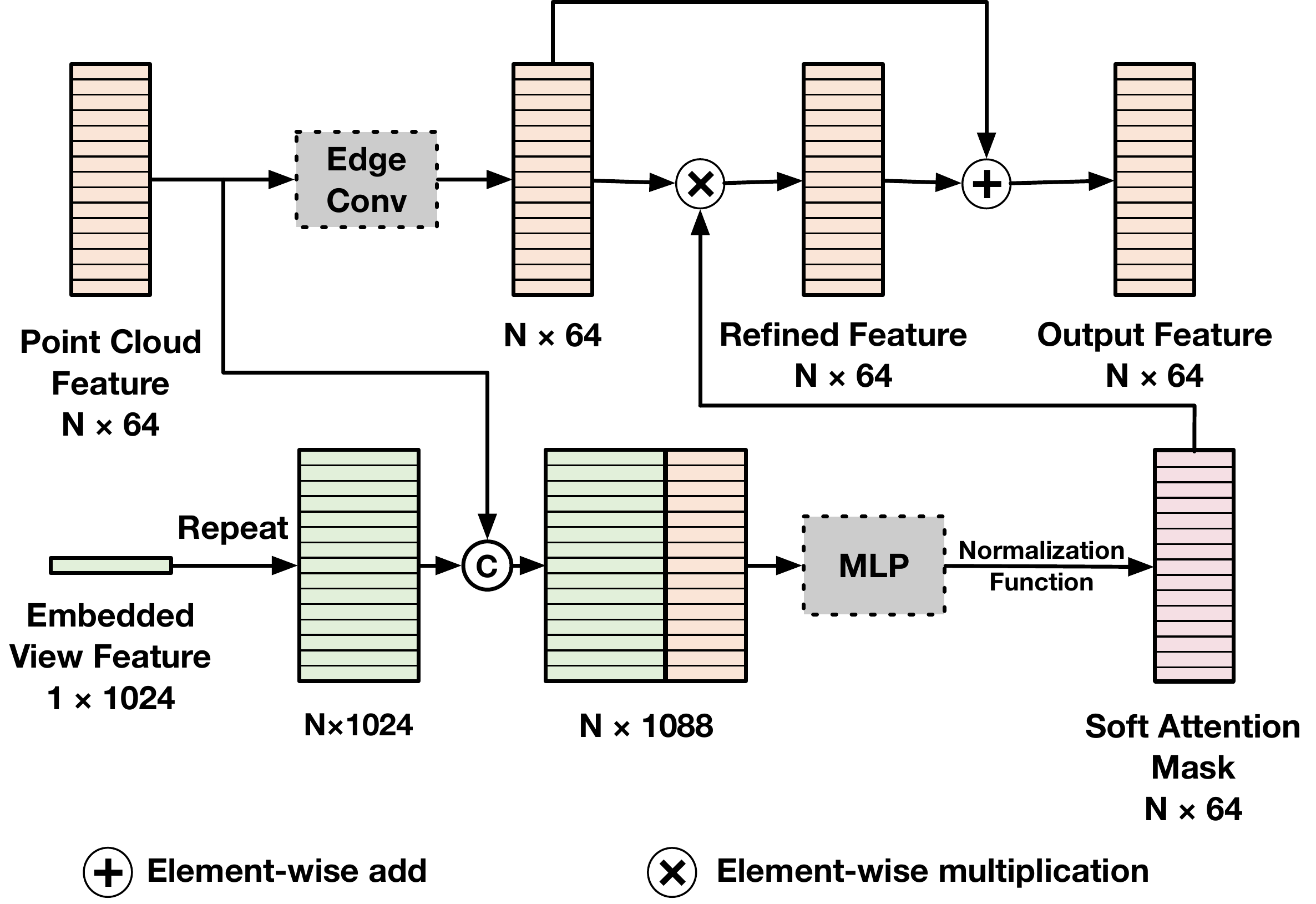}
\end{center}
\vspace{-2mm}
\caption{Attention fusion block. It takes point cloud feature and embedded view feature as input and then repeats the embedded feature N times to concatenate them together, which is followed by a multi-perception layer(MLP) and a normalization function to generate the soft attention mask. Then the mask is applied to the output feature of EdgeConv in a residual way.}
\vspace{-1mm}
\label{fig:attention_block}
\end{figure}

\subsection{Point Cloud Branch and Multi-View Branch}
\textit{Point Cloud Branch}: An $F$-dimensional point cloud with $n$  points could be denoted by $X=\{x_1, \ldots, x_n\}\subseteq \mathbb{R}^F$. Generally, $F=3$ and each point is in 3D coordinates. In the beginning, a 3D spatial transform network is used to compute an affine transformation matrix, which is used in many models \cite{qi2017pointnet,wang2018dynamic} to keep point clouds being invariant to geometric transformations. Then, we use EdgeConv \cite{wang2018dynamic} as the basic layer to be stacked or recurrently applied in networks. EdgeConv first gathers the $k$-nearest neighbors for each point, and then extracts the local edge features of each point by convolution and max pooling, which is a powerful operation to obtain local features.

\textit{Multi-View Branch}: For this branch, we employ the classic MVCNN \cite{su2015multi} as our basic view model. With a set of views for each 3D shape, we feed them separately into a group of CNNs that share the same parameters, and then a view pooling layer is applied to aggregate different view features into a global feature. Containing the high-level information of view representation, this kind of feature would complement point cloud features in our attention embedding fusion.

\subsection{Attention Embedding Fusion}
The method to combine features of two modalities is always a core design for multi-modal task. Previous works often use early fusion \cite{cai2016unified} or late fusion \cite{song2016deep,hoffman2016learning} to handle it. However, early fusion is more suitable for images and videos, and can not be directly applied to the combination of 3D and 2D data. As for the late fusion, it is not efficient enough to exploit both point cloud and view features. More specifically, point cloud based models could effectively capture the local structure features, while the significance of different local features toward object recognition is still not described before aggregating them. As for the view models, the global features are extracted from different views by the well-established CNNs with a view pooling layer. So it is vital to fuse two models in a complementary way. 

Inspired by the recent  advance of attention mechanism in different tasks \cite{wang2017residual,hori2017attention,chen2016attention}, we propose a novel fusion method named attention embedding fusion. Our method mainly focuses on utilizing both the view features in Fig. ~\ref{fig:pipline} and the detailed structure of our attention fusion block is illustrated in Fig. ~\ref{fig:attention_block}. First, the high-level global features obtained by view models are projected into the subspace of point cloud features by an embedding network. Then, we obtain the embedded global feature, that is, an aggregated representation of view information. In addition to direct fusion at the final fully-connected layer, the features are fully leveraged in our attention fusion block. 

\textit{Attention Fusion Block}: Our attention fusion block consists of two paths: fusion mask path and convention path. Given the input point cloud features as a tensor $p$ with shape $n \times c$ ($n$ point with features of $c$ dimensions), EdgeConv is used in convention path to output the local point feature $E(p)$. In fusion mask path, the 1D embedded global feature $v$ of shape $1 \times k$ is concatenated to each input point feature of $p$ by an operation that first repeats embedded feature $n$ times and then concatenates them along each point, where we define as $\phi(\cdot)=\mathbb{R}^{N \times C} \times \mathbb{R}^{1\times K} \to \mathbb{R}^{N \times (C + K)}$. Through this operation and the followed MLP layer, the view global features and point cloud local features are fused together to serve as a relationship descriptor. This descriptor is further quantized by a normalization function $\xi(\cdot)$,
\begin{equation}
  \xi(\cdot)=sigmoid(log(abs(\cdot)))
\end{equation}
which can normalize the output to range $[0,1]$, thus generating a soft attention mask $M(p,v)$. To avoid the output approaching 0 and 1 due to the large input to the sigmoid function, we also add $abs$ and $log$ functions before sigmoid. Then the soft attention mask $M(p,v)$ can be defined as:
\begin{equation}
  M(p,v)=\xi(MLP(\phi(p, v)))
\end{equation}
with the range of [0,1] to represent the significance of different local structures. Then the output of fusion mask path and convention path are merged in a residual way. Directly applying attention mask to convention path could corrupt the original features because of the repeated dot production. So, similar to \cite{wang2017residual}, we employ the residual connection to better utilize the attention masks, and the final output of our attention fusion blocks is defined as 
\begin{equation}
  H(p, v) = E(p)*(1+M(p, v))
\end{equation}
where $E(p) * M(p,v)$ indicates applying the soft attention mask to original features by element-wise multiplication to get the refined feature, which is then added to the original feature $E(p)$. In this way, the attention masks can well describe the relative relationships of different local structures and their contributions to recognize whole 3D objects. Then, our attention fusion blocks act like feature selectors to adaptively augment the meaningful structure feature and to restrain the useless noise feature, making our framework more robust and efficient in feature representation. Additionally, the embedded view features are also concatenated on the last fully-connected layer of point cloud to assist our fusion. Without our attention fusion block, the operation above would degrade into normal late fusion and have limited effect on performance, which we would validate in our subsequent experiments. 

\subsection{Implementation}
\subsubsection{Network Architecture} 
Our framework contains point cloud branch and multi-view branch. For point cloud branch, 1,024 raw points for each object are first processed by a spatial transform network and then sequentially fed into two EdgeConv layers and two attention fusion blocks. In EdgeConv, the number of neighbors $k$ is set to 20 according to \cite{wang2018dynamic}. The embedding network is simplified as a FC layer with satisfactory performance on projecting global features. 
Then the final features combined with multi-view branch are followed by three fully-connected layers to accomplish object recognition. The reason we arrange two attention fusion blocks before the FC layers is that the features become clearer as network going deeper, which are more suitable for attention mechanism. And the first attention block tends to capture mid-level geometric features and the second attention block is more sensitive to high-level geometric features. For multi-view branch, we employ AlexNet, a simple and classic CNN model to validate the efficiency of our framework.
\subsubsection{Training Strategy}
Our framework is trained in an end-to-end fashion. The parameters of CNN in multi-view branch are initialized by the pre-trained MVCNN model. And we adopt an alternative optimization strategy to update our framework. In particular, we first freeze the parameters of multi-view branch and only update our point cloud branch and attention embedding fusion structure for some epochs, and then all the parameters are updated together for some epochs. The reason behind this strategy is that CNNs in multi-view branch have been well pre-trained and point cloud networks are relatively weak in the beginning, which makes our strategy more suitable for the fusion.  
\subsubsection{Classification and retrieval setting}
For the classification task with $C$ categories, the final layer output a vector with shape $C \times 1$, which means the probability that the object belongs to C classes. For retrieval task, the features after out embedding attention fusion are more discriminative to represent object. Therefore,  we employ the feature before the last FC layer to do the retrieval experiment. For two 3D objects $X$ and $Y$, their features are respectively extracted by our PVNet as $x$ and $y$. Then, we use Euclidean distance between two 3D objects for retrieval.

\begin{figure}
\begin{center}
\includegraphics[width=2.7in]{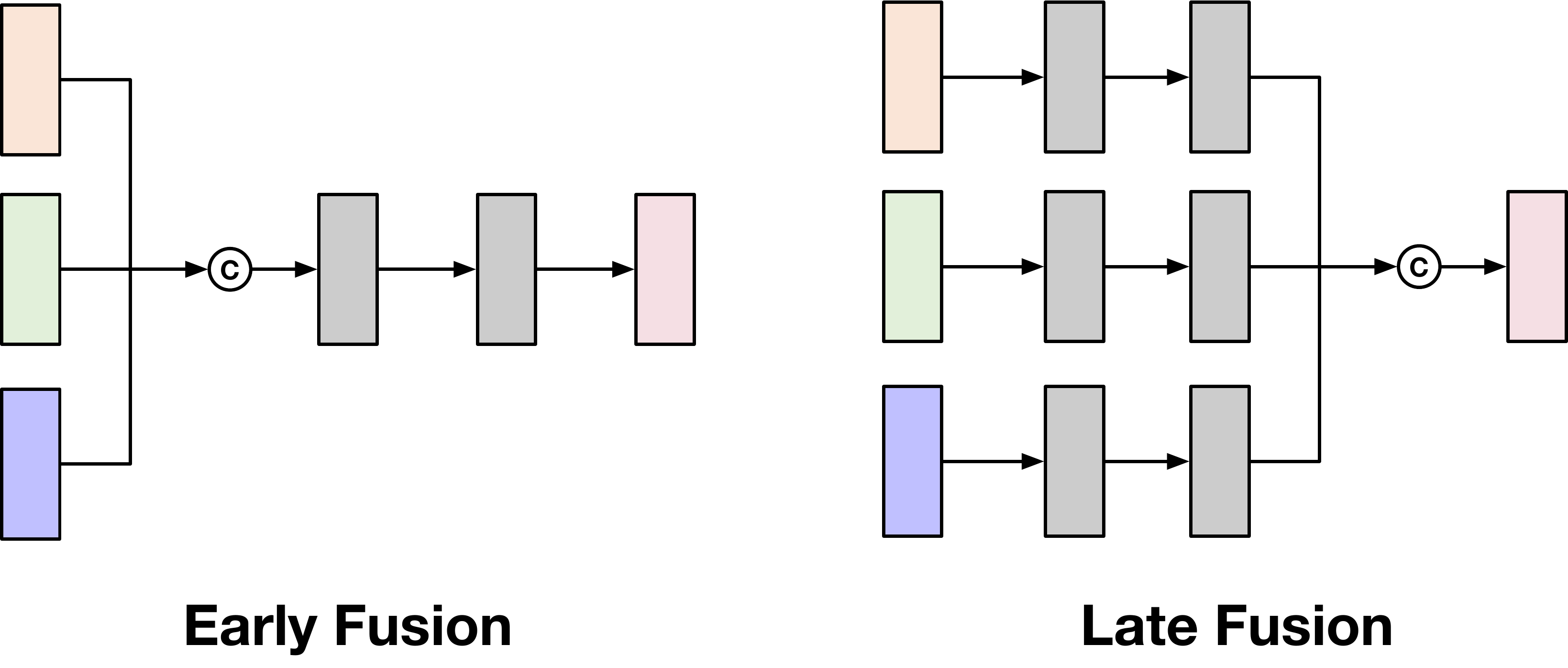}
\end{center}
\caption{Comparison of classical fusion methods. Left: Early fusion method fuses intermediate features. Right: Late fusion method fuses features in the last FC layers.}
\label{fig:fusion}
\end{figure}

\section{Experiments and Discussions}
In this section, we first present the experiments of PVNet on 3D shape classification and retrieval, and also analyze the performance and comparison with the state-of-the-art methods, including both point cloud methods and multi-view methods. We also provide the ablation studies to investigate the proposed embedding attention fusion module. At last, the influence of the number of views and points on the performance of 3D shape classification is investigated for the proposed PVNet.

\begin{table*}[htbp]
\begin{center}
\begin{tabular}{lccccc}
\toprule
\multirow{2}{*}{Method} & \multicolumn{2}{c}{Training Config.} & Data Representation. & Classification  & Retrieval  \\
\cline{2-3}
  & Pre train & Fine tune & \#Number of Views & (Overall Accuracy) & (mAP) \\
\hline
(1)SPH\cite{kazhdan2003rotation} & - & - & - & 68.2$\%$ & 33.3$\%$ \\
(2)LFD\cite{chen2003visual} & - & - & - & 75.5$\%$ & 40.9$\%$ \\
\hline
(3)3D ShapeNets\cite{wu20153d} & ModelNet40 & ModelNet40 & Volumetric & 77.3$\%$ & 49.2$\%$ \\
(4)VoxNet\cite{maturana2015voxnet} & ModelNet40 & ModelNet40 & Volumetric & 83.0$\%$ & - \\
(5)VRN\cite{brock2016generative} & ModelNet40 & ModelNet40 & Volumetric & 91.3$\%$ & - \\
(6)MVCNN-MultiRes\cite{qi2016volumetric} & - & ModelNet40 & Volumetric & 91.4$\%$ & - \\
(7)MVCNN\cite{su2015multi}, 12$\times$ & ImageNet1K & ModelNet40 & 12 Views & 89.9$\%$ & 70.1$\%$ \\
(8)MVCNN\cite{su2015multi}, metric,12$\times$ & ImageNet1K & ModelNet40 & 12 Views & 89.5$\%$ & 80.2$\%$ \\
(9)MVCNN\cite{su2015multi}, 80$\times$ & ImageNet1K & ModelNet40 & 80 Views & 90.1$\%$ & 70.4$\%$ \\
(10)MVCNN\cite{su2015multi}, metric, 80$\times$ & ImageNet1K & ModelNet40 & 80 Views & 90.1$\%$ & 79.5$\%$ \\

(11)MVCNN(GoogLeNet), 12$\times$ & ImageNet1K & ModelNet40 & 12 Views& 92.2$\%$ & 74.1$\%$ \\
(12)MVCNN(GoogLeNet), metric, 12$\times$ & ImageNet1K & ModelNet40 & 12 Views& 92.2$\%$ & 83.0$\%$ \\
(13)FusionNet \cite{hegde2016fusionnet} & ImageNet1K & ModelNet40 & Volumetric and Multi-View& 90.8$\%$ & - \\

(14)PointNet\cite{qi2017pointnet} & - & ModelNet40 & Point Cloud & 89.2$\%$ & - \\
(15)PointNet++\cite{qi2017pointnet++} & - & ModelNet40 & Point Cloud & 90.7$\%$ & - \\
(16)KD-Network\cite{klokov2017escape} & - & ModelNet40 & Point Cloud & 91.8$\%$ & - \\
(17)PointCNN\cite{li2018pointcnn} & - & ModelNet40 & Point Cloud & 91.8$\%$ & - \\
(18)DGCNN\cite{wang2018dynamic} & - & ModelNet40 & Point Cloud & 92.2$\%$ & - \\
\hline

(17)PVNet(AlexNet), 12$\times$ & ImageNet1K & ModelNet40 & Point Cloud and Multi-View & \textbf{93.2}$\%$ & \textbf{89.5}$\%$ \\

\bottomrule
\\
\end{tabular}
\end{center}
\caption{Classification and retrieval results on the ModelNet40 dataset. In experiments, our proposed framework PVNet is compared with state-of-the-art models that use different representations of 3D objects. MVCNN(GoogLeNet) means that GoogLeNet is employed as base architecture for weight-shared CNN in MVCNN. 12$\times$ and 80$\times$ indicate the numbers of views in training. Metric denotes the use of low-rank Mahalanobis metric learning. PVNet(AlexNet) indicates using alexNet as base structure in our multi-view branch and our PVNet can get superior performance over others, especially in the retrieval task.}
\label{tab:experiments}
\end{table*}

\subsection{3D Shape Classification and Retrieval}
To validate the efficiency of the proposed PVNet, 3D shape classification and retrieval experiments have been conducted on the Princeton ModelNet dataset \cite{wu20153d}. Totally, 127,915 3D CAD models from 662 categories are included in the ModelNet dataset. ModelNet40, a common-used subset of ModelNet, containing 12,311 shapes from 40 common categories, is applied in our experiments. We follow the same training and test split setting as in \cite{wu20153d}. For the input of our PVNet, point cloud data and view data are obtained for each 3D model in dataset.

In experiments, we have compared the proposed PVNet with various models based on different representations, including volumetric based models (3D ShapeNets by Wu \textit{et al.} \cite{wu20153d}), hand-craft descriptors for multi-view data (SPH by Kazhdan \textit{et al.} \cite{kazhdan2003rotation} and LFD by Chen \textit{et al.} \cite{chen2003visual}), deep learning models for multi-view data (MVCNN by Su \textit{et al.} \cite{su2015multi} and MVCNN-MultiRes by Qi \textit{et al.} \cite{qi2016volumetric}), and point cloud based models (PointNet by Qi \textit{et al.} \cite{qi2017pointnet}, PointNet++ by Qi \textit{et al.} \cite{qi2017pointnet++}, Kd-Network by Klokov \textit{et al.} \cite{klokov2017escape}, PointCNN by Li \textit{et al.} \cite{li2018pointcnn} and DGCNN by Wang \textit{et al.} \cite{wang2018dynamic}).

\begin{figure}
\begin{center}
\includegraphics[width=2.8in]{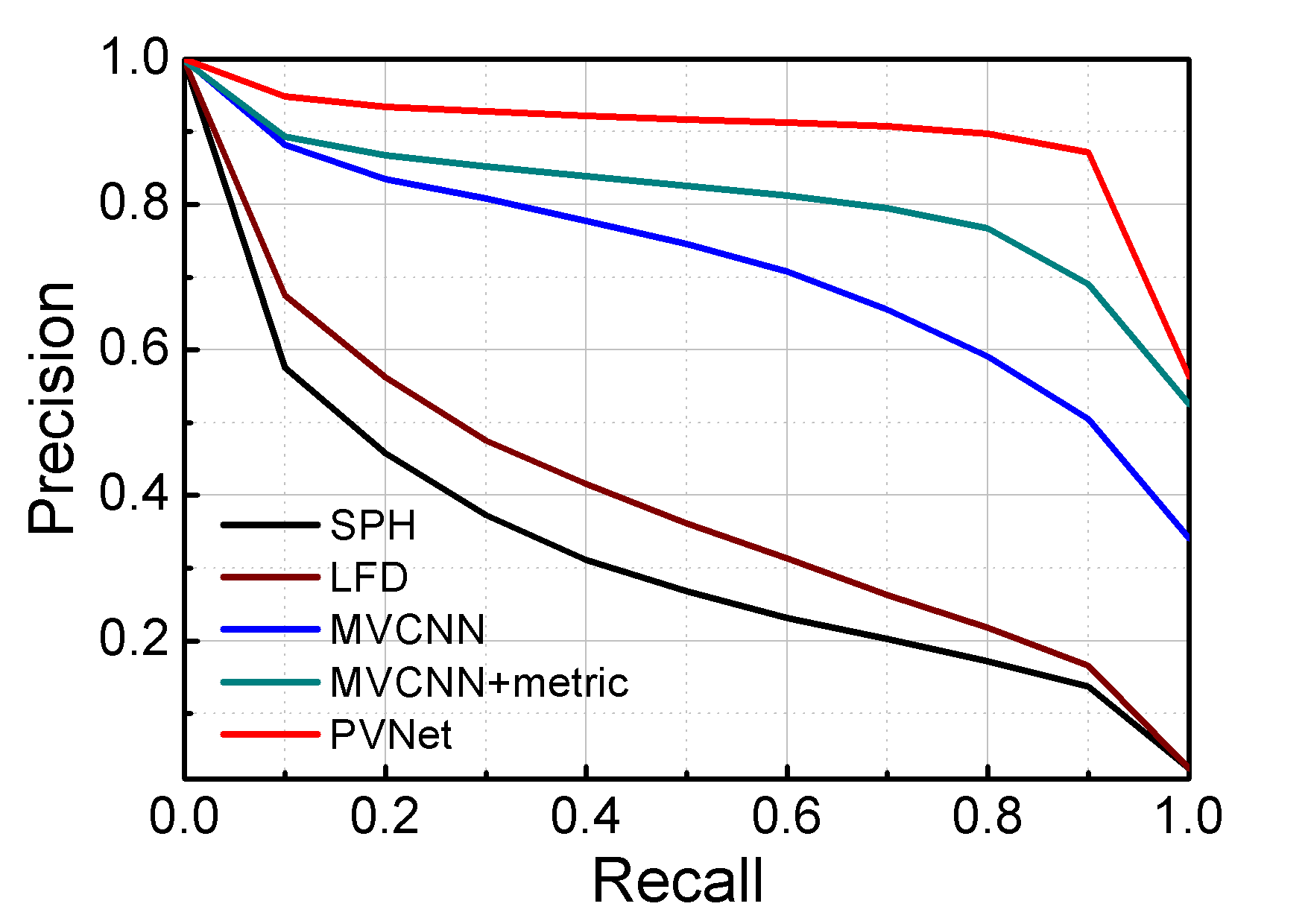}
\end{center}
\vspace{-3mm}
\caption{ Precision-recall curves for our PVNet and other methods on the task of shape retrieval on the ModelNet40 dataset. In these experiments, 12 views and GoogLeNet are used in MVCNN method. The metric denotes using low-rank Mahalanobis metric learning. Our PVNet, without the boost of Mahalanobis metric learning, still significantly outperforms the state-of-the-art and achieves 89.5\% mAP.}
\label{fig:pr}
\end{figure}

In Tab. ~\ref{tab:experiments}, the classification and retrieval results of all compared methods are provided. As shown in the results, our proposed PVNet can achieve the best performance with the classification accuracy of 93.2\% and retrieval mAP of 89.5\%. Compared with the MVCNN using GoogLeNet, our PVNet with AlexNet has gains of 1.0\% and 6.5\% on the classification and the retrieval tasks, respectively. For point cloud based models, our PVNet also outperforms the state-of-the-art point cloud based model DGCNN by 1.0\% in terms of classification accuracy.

In the retrieval task, a low-rank Mahalanobis metric learning is further applied in MVCNNs \cite{su2015multi} to boost the retrieval performance.
While in our proposed PVNet, we directly use the 512-dimensional feature after the first FC layer and achieve an exciting state-of-the-art performance of 89.5\%, which efficiently demonstrates the effectiveness of our PVNet in the 3D shape representation task. The precision-recall curves for retrieval of all compared methods are demonstrated in Fig. ~\ref{fig:pr}. It is obvious that our PVNet, even without Mahalanobis metric learning, can significantly outperform MVCNNs with metric learning and GoogLeNet and all other methods. 

The better performance of our proposed PVNet can be dedicated to the following reasons. First, the proposed method is able to jointly employ both the point cloud and the multi-view data. As point cloud and multi-view have their own advantages and disadvantages for 3D representation, the combination of these two types of 3D data modalities can further enhance the 3D representation performance. Second, the proposed attention embedding fusion module can obtain soft attention masks adaptively to generate the attention-aware features of point cloud models from the embedded global features of view models, which can be effective on 3D shape representation.

\begin{figure}[htpb]
\vspace{0.25cm}
\begin{center}
\subfigure[Overall accuracy]{
\includegraphics[width=1.5in]{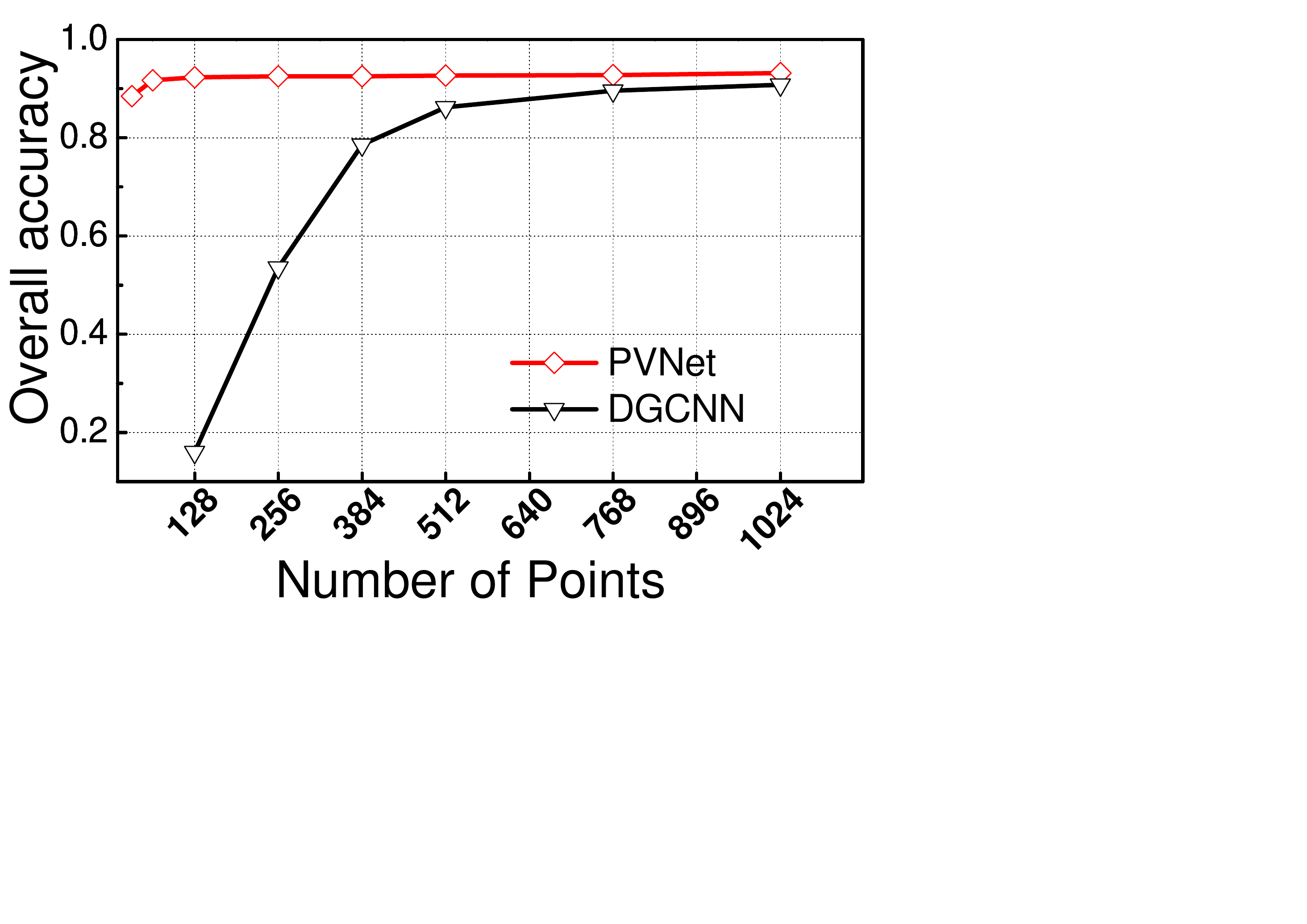}
}
\subfigure[Mean class accuracy]{
\includegraphics[width=1.5in]{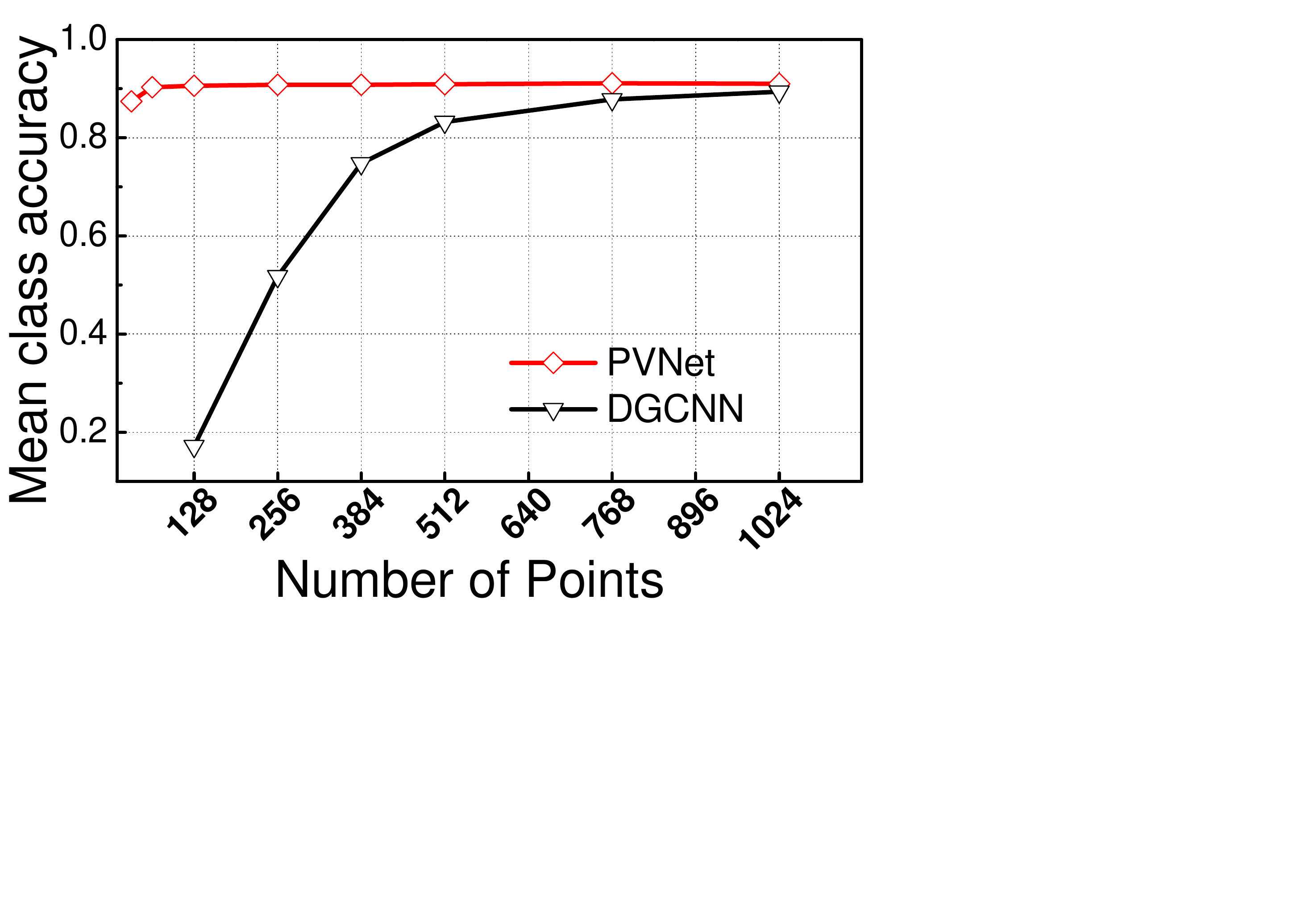}
}
\subfigure[Point clouds with different number of points]{
\includegraphics[width=3in]{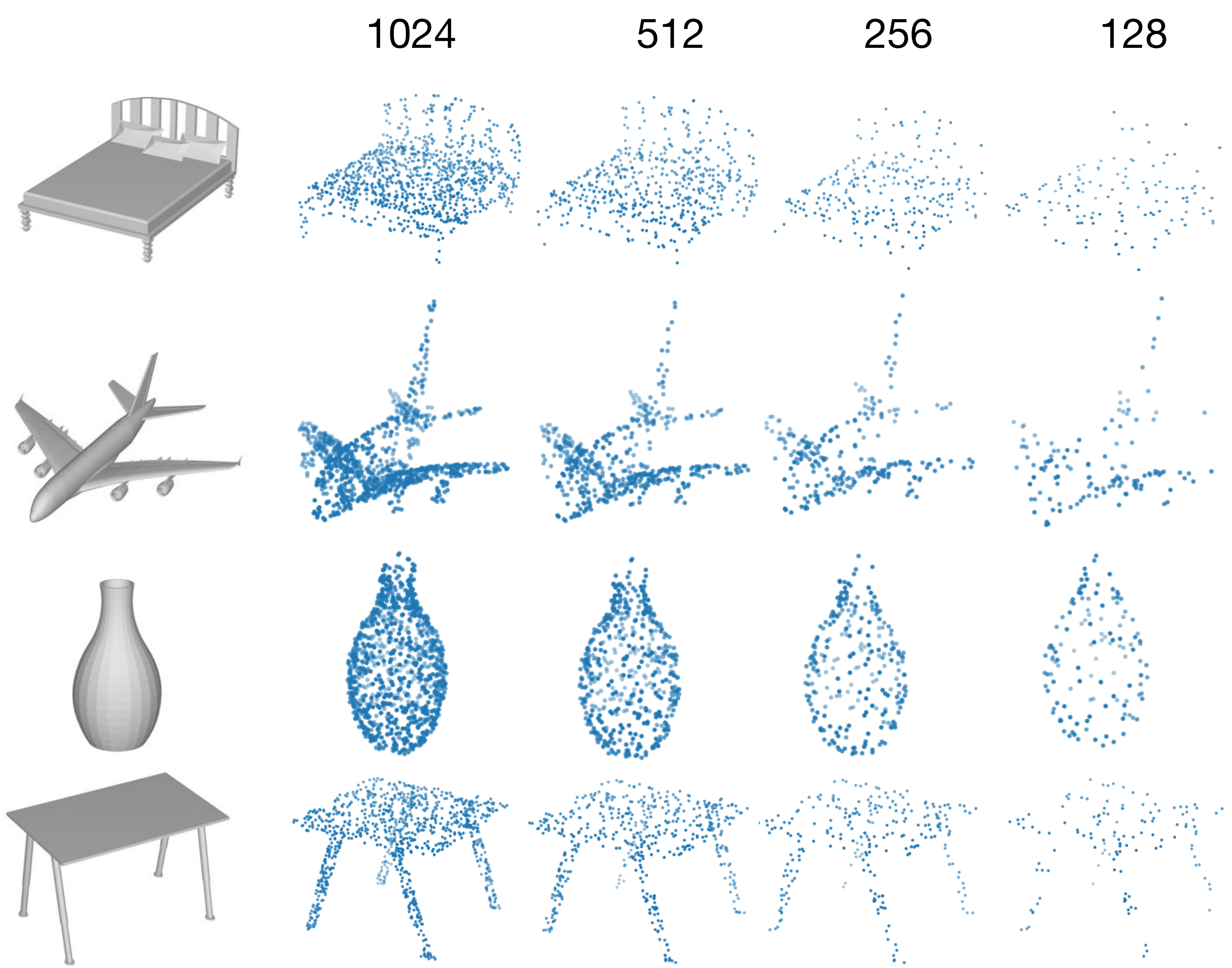}
}
\end{center}
\vspace{-0.3cm}
\caption{The comparison of different numbers of input points. Top: The overall accuracy and mean class accuracy of our PVNet and DGCNN when tested with different numbers of points; Bottom: The examples of the point clouds with different numbers of points.}
\label{fig:pc_miss}
\vspace{-0.2cm}
\end{figure}

\begin{figure*}
\begin{center}
\includegraphics[width=5.4in]{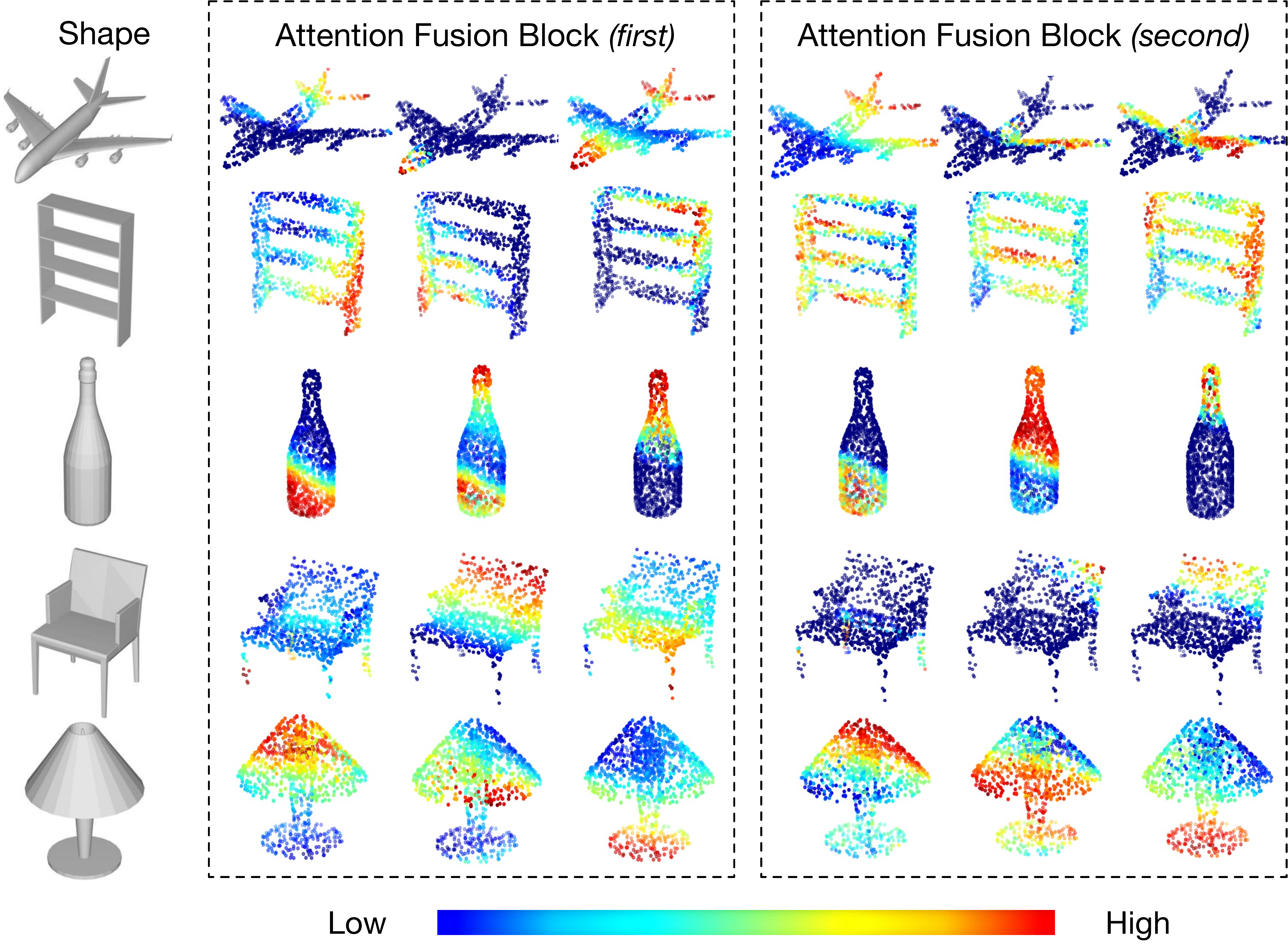}
\end{center}
\vspace{-3mm}
\caption{The visualization of our soft attention masks in PVNet. Left: The 3D shape in our test data. Middle and Right: Soft attention masks of different feature channels in the first and the second attention fusion block. The color from blue to red indicates different weight values ranging from 0 to 1 for each point feature. The high values tend to be assigned in geometric structures that are critical for recognition, such as the wings of planes. And the attention masks in second block are more sensitive to high-level feature such as the layer structures in bookshelves.}
\label{fig:mask}
\end{figure*}

\subsection{Ablation Studies}

In our PVNet, the proposed embedding attention fusion plays an important role. In this sub-section, we further deeply investigate the embedding attention fusion module.

In multimodal task, early fusion and late fusion are two typical fusion methods, as shown in Fig. ~\ref{fig:fusion}. For point cloud and multi-view models, early fusion is intractable due to the different dimensions of two representations and their shallow features. The proposed method without the attention fusion block and the embedding network can degrade into a traditional late fusion scheme.

\begin{table}
  \caption{Effectiveness of different components of our framework on the classification task.}
  \vspace{-3mm}
  \label{tab:ablation}
  \begin{tabular}{ccc}
    \toprule
    Models&Mean Class Accuracy&Overall Accuracy\\
    \midrule
    Point Cloud Model& 90.2\% & 92.2\%\\
    Multi-View Model & 87.6\% & 89.9\%\\
    Late Fusion & 90.8\% & 92.5\%\\
    Ours & \textbf{91.0\%}& \textbf{93.2\%}\\
  \bottomrule
\end{tabular}
\end{table}

We then evaluate the performance of the proposed method using different combinations of the components and demonstrate the results in Table ~\ref{tab:ablation}. In this table, "Point Cloud Model" denotes only the point cloud branch of our method is used,  indicating that one spatial transform net and four EdgeConv layer and following pooling and FC layers are applied. "Multi-View Model" denotes only the multi-view branch is used. "Late Fusion" denotes that the point cloud global feature and the multi-view global feature are directly concatenated in a late fusion way. "Ours" denotes the proposed embedding attention fusion method. As shown in the comparison, we observe that our method can outperform not only the baseline point cloud model and multi-view model by a large margin, but also outperform the late fusion model by 0.7\% in overall accuracy and 0.2\% in mean class accuracy in the classification task. That is dedicated to our well-designed framework of jointly using both the point cloud data and the multi-view data. Our framework makes it possible to employ the global features of multi-view models to describe the correlations of different geometric properties extracted by point cloud models and their significance for shape representation, which could be quantified into a soft attention mask. Then, more discriminative geometric features can be captured by applying the soft attention mask in a residual way.

To better illustrate the intuition behind our attention fusion block, we visualize the soft attention masks in Fig. \ref{fig:mask}. We find the masks in each channel focus on different geometric features, which are important structures for 3D shape recognition. It means that our masks can learn to assign relative low weights for the unimportant point features for recognition and high weights for more discriminative features. And thus, the masks role like feature selectors to enhance good features and discard useless features like noise, which help to refine features and bring the performance improvement.

\subsection{On The Number of Views and Points}
In this sub-section, we mainly focus on another critical issue about the robustness of our framework. In practice, dealing with missing data is a common situation, where only parts of the data can be used. Under such circumstances, it is important to investigate the robustness and generalization ability of the model in different data situations.

First, we keep the point cloud data as complete and vary the number of views. The number of employed views is selected as 4, 8 and 12, respectively, and we compare the proposed method with the multi-view method, i.e., MVCNN. The classification results are shown in Tab. \ref{tab:view_miss}. As shown in the results, we can observe that the missing data will reduce the performance for both the proposed method and MVCNN, and the proposed method can still achieve better performance compared with MVCNN.

\begin{table}
  \caption{The comparison of different numbers of views. 12 views are used to train models and we separately employ 4, 8, 12 views as our test data that are captured by cameras with 90$^\circ$, 45$^\circ$, 30$^\circ$ interval. Top: The performance of MVCNN; Bottom: The performance of our PVNet.}
  \vspace{-3mm}
  \label{tab:view_miss}
  \begin{tabular}{ccc}
    \toprule
    Number of Views&Mean Class Accuracy&Overall Accuracy\\
    \midrule
    4& 82.5 \% & 84.6\%\\
    8& 87.2\% & 89.0\%\\
    12& 87.6\% & 89.9\%\\
    \midrule
    4& 85.9 \% & 87.2\%\\
    8& 90.0\% & 91.8\%\\
    12& 91.0\% & 93.2\%\\
  \bottomrule
\end{tabular}
\end{table}


Second, we keep the multi-view data as complete and vary the number of input points. The number of employed points is selected as 128, 256, 384, 512, 768, 1,024, respectively, and we compare the proposed method with the current state-of-the-art point cloud based method, i.e., DGCNN. The classification results are shown in Fig. \ref{fig:pc_miss}. As shown in the results, we can observe that the missing data will reduce the performance significantly for DGCNN. For instance, when only 128 points are available, the overall accuracy for DGCNN is lower than 20\%. For the proposed PVNet, it is stable with respect to the missing point cloud data. It only drops about 1\% and 0.5\% when there are only 128 points used for testing in terms of overall accuracy and mean class accuracy, respectively. While when the number of points drops significantly, such as only 32 points are available, our framework can still achieve relatively good performance, i.e, 88.5\% for overall accuracy. This stable performance comes from the compensation of corresponding multi-view data, which effectively reduces the influence of missing point cloud data.


The experimental results and comparisons can demonstrate the robustness of our proposed PVNet on missing data. This is attributed to our complementary fusion method, which can still keep performance with missing data. When the data from one modality is corrupt, the features from the other modality help to compensate.

\section{Conclusion}
In this paper, we propose the first convolutional network, i.e., PVNet, which can jointly employ point cloud data and multi-view data for 3D shape recognition. In our framework, the embedding attention fusion is introduced to employ the global view feature from the multi-view branch to help to portray the correlation and significance of different geometric properties extracted by the point cloud branch. Then, this discriminative information is quantified into a soft attention mask, which helps to further capture attention-aware features. Different from previous models, our method can efficiently explore the complementary relation between point cloud and multi-view as two representations for 3D shapes. The effectiveness of our proposed framework has been demonstrated by experimental results and comparisons with the state-of-the-art models on the ModelNet40 dataset. We have also investigated the effect of different components of our model, as well as the influence of different numbers of views and point clouds for 3D shape representation, which demonstrates the robustness of our framework.



\begin{acks}
    This work was supported by National Key R$\&$D Program of China (Grant No. 2017YFC0113000), and National Natural Science Funds of China (U1701262, 61671267, U1705262, 61772443, 61572410).
\end{acks}

\bibliographystyle{ACM-Reference-Format}
\balance
\bibliography{sample-bibliography}

\end{document}